\documentclass[letterpaper, 10 pt, conference]{ieeeconf}  
\usepackage{amssymb}
 \usepackage[pdftex]{graphicx}
 \usepackage{amsmath}
\usepackage{balance}
\usepackage[linesnumbered,ruled]{algorithm2e}

\newcommand{\argmin}{\mathop{\rm arg~min}\limits}

\IEEEoverridecommandlockouts                              

\title{\LARGE \bf
MultiCruise: Eco-Lane Selection Strategy with\\Eco-Cruise Control for Connected and Automated Vehicles
}

\author{Shunsuke Aoki$^{1,2}$, Lung En Jan$^{2}$, Junfeng Zhao$^{3}$, Anand Bhat$^{2}$,\\
Chen-Fang Chang$^{3}$, Ragunathan (Raj) Rajkumar$^{2}$
\thanks{$^{1}$National Institute of Informatics, Japan
        }%
\thanks{$^{2}$Electrical \& Computer Engineering, Carnegie Mellon University
        }%
\thanks{$^{3}$General Motors Company
        }%
}

\begin{document}

\maketitle
\thispagestyle{empty}
\pagestyle{empty}

\begin{abstract}


Connected and Automated Vehicles (CAVs) have real-time information from the surrounding environment by using local on-board sensors, V2X (Vehicle-to-Everything) communications, pre-loaded vehicle-specific lookup tables, and map database.
CAVs are capable of improving energy efficiency by incorporating these information.
In particular, Eco-Cruise and Eco-Lane Selection on highways and/or motorways have immense potential to save energy, because there are generally fewer traffic controllers and the vehicles keep moving in general.
In this paper, we present a cooperative and energy-efficient lane-selection strategy named {\it MultiCruise}, where each CAV selects one among multiple candidate lanes that allows the most energy-efficient travel.
MultiCruise incorporates an Eco-Cruise component to select the most energy-efficient lane.
The Eco-Cruise component calculates the driving parameters and prospective energy consumption of the ego vehicle for each candidate lane, and the Eco-Lane Selection component uses these values.
As a result, MultiCruise can account for multiple data sources, such as the road curvature and the surrounding vehicles' velocities and accelerations.
The eco-autonomous driving strategy, MultiCruise, is tested, designed and verified by using a co-simulation test platform that includes autonomous driving software and realistic road networks to study the performance under realistic driving conditions.
Our experimental evaluations show that our eco-autonomous MultiCruise saves up to $8.5 \%$ fuel consumption.

%

\end{abstract}

\section{INTRODUCTION}


\vspace{1mm}

Fuel efficiency is one of the most significant factors for vehicles.
In fact, according to the U.S. Energy Information Administration (EIA) \cite{US_EIA}, nearly 143 billion gallons of motor gasoline were consumed in 2018 in the United States.
Also, $CO_2$ emissions due to motor gasoline and diesel fuel consumption comprise over $30 \%$ of total U.S. energy-related $CO_2$ emissions.

\vspace{1mm}

Connected and Automated Vehicles (CAVs) have immense potential to improve fuel economy.
Even though vehicle automation alone is unlikely to have significant impacts on fuel consumption \cite{wadud2016help}, CAVs can effectively improve fuel efficiency by using intersection connectivity \cite{aoki2017configurable, aoki2019DSIP}, platooning \cite{tsugawa2016review}, eco-driving strategies \cite{peter2020ecodriving} and other novel applications \cite{brown2014analysis}.
With using the peripheral technologies, such as vehicular communications and/or edge computing, CAVs are capable of improving energy efficiency by incorporating a variety of information.

\vspace{1mm}

In particular, Eco-Lane Selection and Eco-Cruise on highways/motorways are promising eco-driving strategies, because there are generally fewer traffic controllers and the vehicles keep moving in general.
In fact, human drivers change the lane when there is a slower lead vehicle and when the neighboring lane has sufficient space, in order to shorten their travel times and save fuel.
The Eco-Cruise maneuver adjusts vehicle speed and acceleration based on traffic, speed limit changes, road features and navigational maneuvers \cite{peter2020ecodriving}.
Eco-Lane Selection determines the lane to drive in and the timing to change the lane, accounting for road structures and traffic.
In these eco-autonomous driving strategies, CAVs incorporate static/dynamic data sources, including the on-board sensors, V2X (Vehicle-to-Everything) communications, pre-loaded vehicle-related data, and map database.
Although V2V (Vehicle-to-Vehicle) communications are not necessary to save fuel, cooperation by V2V communications is at least beneficial for road safety \cite{lehmann2018generic, hess2019negotiation}.

\begin{figure}[!b]
\centering
\includegraphics[width=6.75cm]{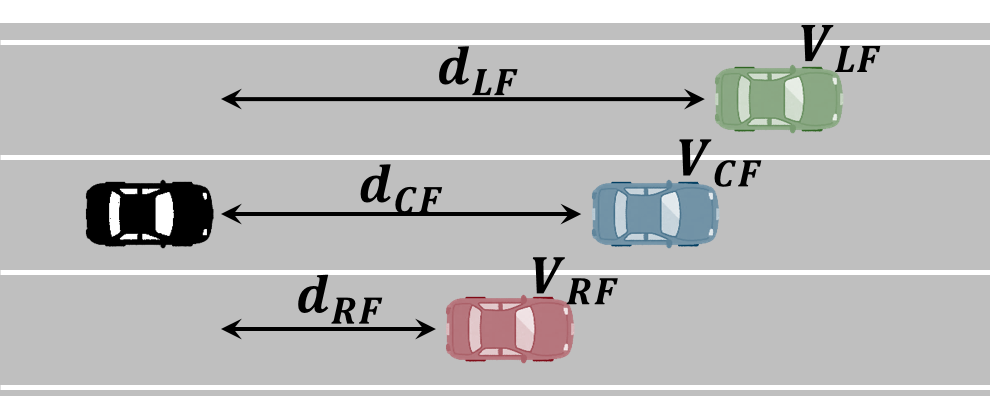}
\caption{\textit{MultiCruise} for Energy-Efficient Lane Selection Strategy.}
\label{fig:EcoLaneselection}
\end{figure}

\vspace{1mm}

In this paper, we present a cooperative and energy-efficient lane-selection strategy named {\it MultiCruise} in which each CAV selects the optimal lane and optimal target vehicle to follow among multiple candidates by working with the Eco-Cruise component.
In MultiCruise, as shown in Figure \ref{fig:EcoLaneselection}, the Eco-Cruise component \cite{peter2020ecodriving} calculates multiple sets of the driving parameters of the ego vehicle, and then, the Eco-Lane Selection component determines the lane to drive for saving fuel consumption while ensuring road safety.
MultiCruise accounts for multiple data sources, such as the road curvature, grade, and the surrounding vehicles' velocities and accelerations.
MultiCruise calculates two costs to determine the goal behavior: (i) the cost for staying in the current lane and (ii) the cost for changing the lane and driving the target lane.
The cost function includes fuel costs and penalties.
The fuel costs are for saving fuel, while penalties are for enabling cooperative and collaborative driving to improve road safety during lane-changing maneuvers.
By using these penalties, MultiCruise is able to avoid selfish back-to-back lane-changing maneuvers, which may affect the road safety for the surrounding traffic.
Since CAVs are safety-critical and life-critical applications, MultiCruise never allows a lane-changing maneuver when road safety can be compromised.
Hence, MultiCruise improves fuel economy while satisfying safety constraints.

The primary contributions presented in this paper are as follows:

\vspace{1mm}
\begin{itemize}
\item We present an Eco-Lane Selection strategy named \textit{MultiCruise} that selects the optimal lane and optimal target vehicle to follow to save fuel consumption.
\item We model and develop a cooperative mechanism for a safe lane-changing maneuver that avoids selfish back-to-back lane-changing maneuvers.
\item We test and verify MultiCruise on a co-simulation platform \cite{aoki2020inforich} that includes autonomous driving software and realistic road networks.
\end{itemize}
\vspace{1mm}

The remainder of this paper is organized as follows.
Section I\hspace{-.1em}I reviews previous work related to our research.
Next, Section I\hspace{-.1em}I\hspace{-.1em}I describes the lane-changing maneuvers focused in this paper.
Section I\hspace{-.1em}V presents our eco-lane selection strategies.
Section V gives the implementation and evaluation of our eco-driving strategies.
Finally, Section V\hspace{-.1em}I presents our summary and future work.

\vspace{0.5mm}


\section{RELATED WORK}

The advent of Connected and Automated Vehicles (CAVs) \cite{urmson2008autonomous} is a great opportunity to improve not only safety but also vehicle fuel economy.
In fact, ARPA-E runs NEXTCAR program (Next-Generation Energy Technologies for Connected and Automated On-Road Vehicles) that aims to leverage the CAV technologies to further improve the energy efficiency \cite{aoki2020inforich, olin2019reducing, oncken2020connected}.
Vehicle automation alone is unlikely to have significant impacts on fuel savings \cite{wadud2016help}, but there are multiple promising technologies for fuel savings, such as vehicle platooning, intersection connectivity, traffic management and eco-autonomous driving strategies.

In particular, eco-autonomous driving strategies can effectively improve fuel efficiency and they include four eco-driving applications: Eco-Approach, Eco-Departure \cite{barth2019ead, shao2020eco}, Eco-Cruise \cite{peter2020ecodriving, jing2016design, firoozi2019safe} and Eco-Lane Selection \cite{aoki2020towards}.
Eco-Approach is defined as the fuel-efficient vehicle/powertrain operation that applies smooth deceleration to bring a vehicle to stop in fuel-friendly fashion.
The Eco-Departure maneuver should conduct a smooth and less-aggressive acceleration profile to reach the target cruise speed. The target speed and the distance to reach the speed typically depend on the traffic preview information that are available from on-board sensors and vehicular communications. Eco-Approach and Eco-Departure are closely related to road intersections and traffic controllers \cite{barth2019ead}, especially signalized intersections.
For example, by using the SPaT (Signal Phase and Timing) information via vehicular communications, each connected vehicle can save fuel.

In addition, Eco-Cruise is the vehicle velocity control strategy to save the fuel consumption.
One of the promising techniques for the cruise control is a Model Predictive Control (MPC), and Jing et al. \cite{jing2016design} used a Finite State Machine (FSM) to narrow down the search space for each MPC step to be used for real-time applications.
Also, Firoozi et al \cite{firoozi2019safe} presented a Eco-Cruise component that incorporates the road grade information and V2V communications.
In \cite{peter2020ecodriving}, Eco-Cruise is designed and developed to determine the driving parameters of the ego vehicle by using the given lane and given target vehicle at the moment. The Eco-Cruise component accounts for road grade, curvature and surrounding traffic, and we use this component for MultiCruise.

Eco-Lane Selection is also a significant application for fuel savings.
FWHA's Next Generation SIMulation (NGSIM) program \cite{NGSIM2006} has defined ``target lane'' concept and also has introduced Freeway Lane Selection (FLS) algorithm, for modeling a lane-changing maneuver on highways and/or motorways.
For connected vehicles, Jin et al \cite{jin2014improving}, Kang et al. \cite{kang2019development} and Tian et al. \cite{tian2018connected} proposed optimal lane-selection algorithms for traffic operations and management on highway segments. Since these works relied on V2I (Vehicle-to-Infrastructure) communications and segment-based operations, the approaches were not scalable. Also, there are multiple on-going studies for lane-changing and merging maneuvers \cite{aoki2018dynamic, aoki2017merging} for autonomous vehicles.
Unlike our work presented in this paper, these works just focused on the throughput and never discussed fuel efficiency.



%
%

\section{MANDATORY AND DISCRETIONARY LANE-CHANGING MANEUVERS}

In this section, we present a review of lane-changing maneuvers on highways.
Lane-changing maneuvers have been originally studied for human-driven vehicles \cite{jin2014improving, goswami2007gap, zhao2017analysis}.
These maneuvers can be categorized into two classes \cite{toledo2003modeling}: (i) \textbf{Mandatory lane-changing} and (ii) \textbf{Discretionary lane-changing}.
First, mandatory lane-changing maneuvers are conducted for vehicle navigation.
When a vehicle has to change its lane to drive, for lane reduction or for exiting from highways, the lane-changing maneuver is classified into mandatory one.
For such mandatory lane-changing maneuvers, when the vehicle cannot complete its maneuver until the deadline, it may have to re-calculate its route by using the navigation system.
On the other hand, discretionary lane-changing maneuvers are for secondary purposes, such as energy efficiency, traffic throughput and/or passenger comfort.
A vehicle does not need to complete the discretionary lane-changing maneuver for navigation purposes.
Several researchers have studied the lane-changing maneuvers for autonomous driving \cite{dong2018continuous, krasowski2020safe}, but they do not focus on such features \cite{dong2018continuous, krasowski2020safe}.


\vspace{1mm}

In this paper, we focus on discretionary lane-changing maneuvers, because our primary objective is saving fuel consumption on highways and motorways.
Here, MultiCruise for CAVs is the lane-selection component for saving energy and it is not for vehicle navigation and route planning. Therefore, each CAV only uses MultiCruise to change its lane when it can follow the route provided by its navigation system.

\section{MULTICRUISE: AN ECO-LANE SELECTION STRATEGY}

We now present MultiCruise as an energy-efficient lane-selection strategy.
MultiCruise consists of $2$ steps to determine the lane to drive for saving fuel consumption: (i) \textit{Calculating cost for each potential target} and (ii) \textit{Determining goal behaviors}.
In this section, we first present the cost function, and secondly propose a decision-making policy.
In addition, to show the safety and practicality of MultiCruise, we discuss cooperative and safe mechanisms for MultiCruise.

Each CAV uses MultiCruise by incorporating various data sources, including the on-board sensors, V2X (Vehicle-to-Everything) communications, pre-loaded vehicle-specific lookup tables, and map database.
Since each vehicle determines its behaviors in a distributed manner, our MultiCruise is designed to be used both around human-driven vehicles and around CAVs seamlessly.


\subsection{Cost and Penalty: For Multiple Targets}

Each CAV uses MultiCruise to calculate three Eco-Driving costs to determine whether it should change the driving lane for saving fuel consumption. These three Eco-Driving costs are captured in Eq. (\ref{eq:three_cost}).
Here, we calculate the driving costs on the left neighboring lane $C_{LF}$, current lane $C_{CF}$, and right neighboring lane $C_{RF}$, by using the cruise cost $\Gamma$ introduced and discussed in \cite{peter2020ecodriving} and the penalty $P$ for safety purposes.
The penalty $P$ is presented in Eq. (\ref{eq:penalty_function}).

\begin{eqnarray}
\left\{ \begin{array}{lll}
C_{LF} &= & \Gamma(a_{Ego}, v_{Ego}, v_{LF}, d_{LF}) + P(t_\Delta, v_{Ego})\\
C_{CF} &= & \Gamma(a_{Ego}, v_{Ego}, v_{CF}, d_{CF}) \\
C_{RF} &= & \Gamma(a_{Ego}, v_{Ego}, v_{RF}, d_{RF}) + P(t_\Delta, v_{Ego})\\
\end{array} \right.
\label{eq:three_cost}
\end{eqnarray}

\begin{equation}
P(t_\Delta, v_{Ego}) = s \cdot \frac{v_{Ego}}{t_\Delta}
\label{eq:penalty_function}
\end{equation}

Here, $a_{Ego}$ and $v_{Ego}$ are the ego vehicle's acceleration and speed, respectively.
$d_{LF}$, $d_{CF}$ and $d_{RF}$ represent the inter-vehicle distances to the lead vehicle on the left lane, current lane, and right lane, respectively, as shown in Figure \ref{fig:EcoLaneselection}.
Also, $v_{LF}$, $v_{CF}$ and $v_{RF}$ are the lead vehicles' speeds on the left lane, current lane, and right lane, respectively.
The cruise cost $\Gamma$ \cite{peter2020ecodriving} is calculated in the Eco-Cruise component and it includes (i) fuel cost, (ii) progress cost, and (iii) comfort cost.
The fuel cost is calculated by using the vehicle-specific fuel table, to get the expected fuel consumption from the vehicle velocity and acceleration.
The progress cost is designed to penalize when the ego vehicle is too slow and does not make progress over time.
The comfort cost accounts for vehicle jerk, in order to provide comfortable driving to the human passengers.
A more detailed discussion on jerk is presented in \cite{peter2020ecodriving}.

In addition, the penalty $P$ in Eqs. (\ref{eq:three_cost}) and (\ref{eq:penalty_function}) is designed for avoiding back-to-back and aggressive lane-changing maneuvers.
Here, $t_\Delta$ describes the elapsed time from the last lane-changing maneuver. $s$ is a scaling factor.
By using this mechanism, the penalty becomes very large value right after the lane-changing maneuver is completed, and consequently, MultiCruise does not encourage the lane-changing maneuver at this moment.
The vehicle may be able to reduce fuel consumption by conducting back-to-back lane-changing maneuvers, but this behavior compromises road safety \cite{li2006realistic} and we do not encourage this behavior in MultiCruise.

Each CAV uses these $3$ Eco-Driving costs, $C_{LF}$, $C_{CF}$ and $C_{RF}$, to determine its behavior. We will propose and discuss the decision-making policy next.

\begin{figure}[!t]
\centering
\includegraphics[width=8.55cm]{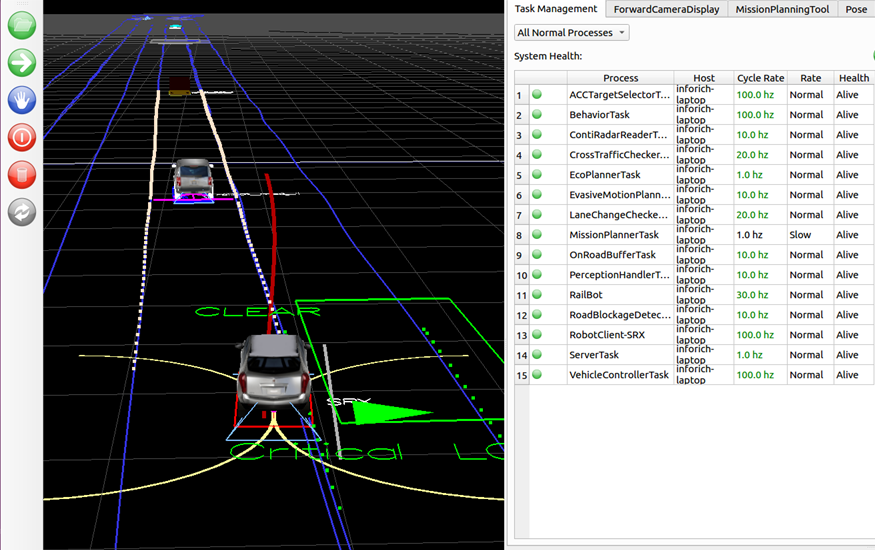}
\caption{Space Checker for Lane-Changing Maneuvers.}
\label{fig:Lane_selection_target}
\end{figure}

\subsection{Decision-making Policy}

MultiCruise compares the $3$ Eco-Driving costs and selects one lane to save fuel consumption.
To guarantee road safety under the practical CAV systems, MultiCruise has to satisfy $2$ requirements: (i) MultiCruise does not overturn the decision of the navigation system, and (ii) MultiCruise enables a safe lane-changing maneuver.
For the first requirement, we use a threshold distance $D_{Thr}$ and a distance for the navigation $D_{NAV}$.
$D_{NAV}$ represents the distance to the end of the road segment and/or ramp-off.
When $D_{NAV}$ is smaller than $D_{Thr}$, each vehicle has to consider the mandatory lane-changing maneuver for its navigation purpose.
For the second requirement, after determining the target lane $\theta$, each vehicle has to confirm that the target lane has the sufficient space to complete its lane-changing maneuver by using the sensor-based perception. Since the lane-changing maneuvers may lead vehicle accidents and/or collisions, each vehicle can start the lane-changing maneuver only when safety is guaranteed.

The decision-making algorithm for MultiCruise is captured in Algorithm \ref{algorithm:decition-making}.
Here, we define and use $C_{Thr}$ as a threshold value for the cost, in order to avoid unnecessary lane-changing maneuvers, which may result only in trivial benefits.
Since the lane-changing maneuvers may disturb the entire traffic flow, MultiCruise does not encourage the aggressive and/or unnecessary lane-changing maneuvers.

\begin{algorithm}[b!]
    \If{$D_{NAV} > D_{Thr}$}
    {
      Calculate $C_{LF}$, $C_{CF}$, $C_{RF}$\;
      \If{$C_{CF}>C_{Thr}$}
      {
       target lane $\theta = \argmin_{\alpha \in \{LF, LC, LR\}} C_{\alpha}$\;
      }
    }
\caption{Decision-making algorithm for MultiCruise}
\label{algorithm:decition-making}
\end{algorithm}

\subsection{Safety for Lane-changing Maneuver}

Although the primary objective of MultiCruise is saving fuel consumption, to enhance the road safety for lane-changing maneuvers, MultiCruise includes $2$ mechanisms: (i) Passed time $t_\Delta$ and (ii) Space checker.

First, $t_\Delta$ represents the passed time from the last lane-changing maneuver.
When the value $t_\Delta$ is small, the penalty $P$ becomes a large value and MultiCruise discourages the lane-changing maneuver at this moment.
Therefore, after each vehicle changes its lane to drive, it stays in that lane for a certain time period.
Since back-to-back lane-changing maneuvers disturb the surrounding traffic and may lead to collisions, this cooperative mechanism enhances road safety.

Secondly, when each vehicle uses MultiCruise, it always confirms the space on the target lane, as shown in Figure \ref{fig:Lane_selection_target}, by using its space checker.
In Figure \ref{fig:Lane_selection_target}, the green box represents the area confirmed by the space checker. When there is a moving obstacle within the area, MultiCruise stops the lane-changing maneuver, because it cannot guarantee road safety.
MultiCruise uses the space checker after the target lane $\theta$ is selected by the decision-making algorithm described in Algorithm \ref{algorithm:decition-making}.

By using these mechanisms, MultiCruise improves road safety while enhancing fuel efficiency on highways and/or motorways.


\begin{figure}[!b]
\centering
\includegraphics[width=8.55cm]{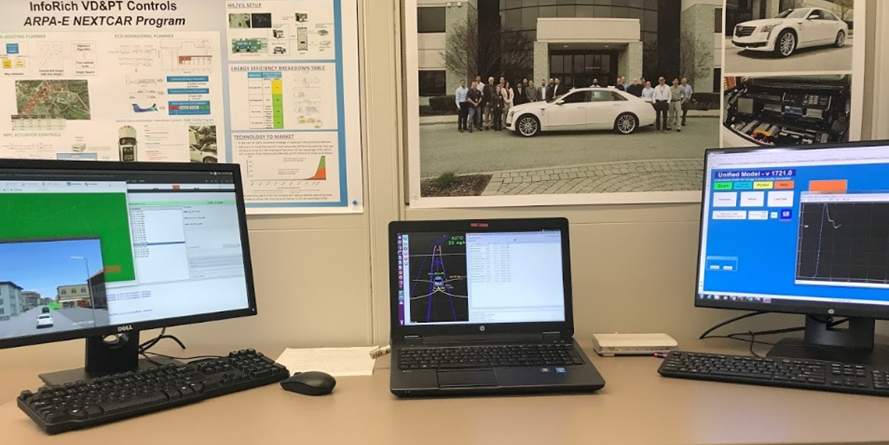}
\caption{Co-simulation Hardware Setup.}
\label{fig:harware_setup}
\end{figure}

\begin{figure}[!b]
  \begin{center}
    \begin{tabular}{c}
      \begin{minipage}{0.45\hsize}
        \begin{center}
          \includegraphics[width=4.15cm]{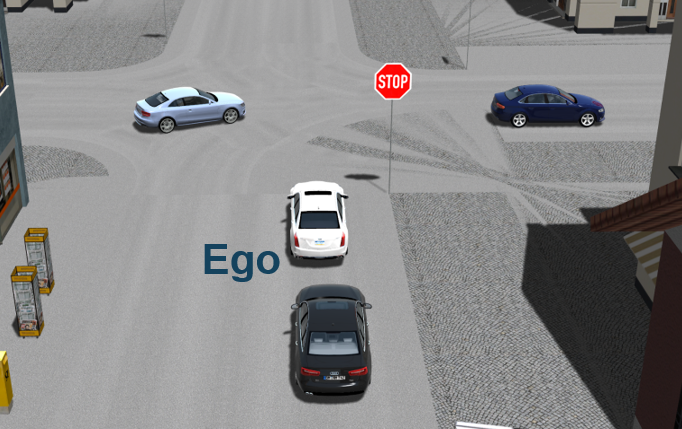}
	\hspace{7.2cm} (a) Moving Obstacles \\ in Traffic Simulator.
        \label{fig:Synchro_2size}
        \end{center}
      \end{minipage}
      \begin{minipage}{0.55\hsize}
        \begin{center}
          \includegraphics[width=4.15cm]{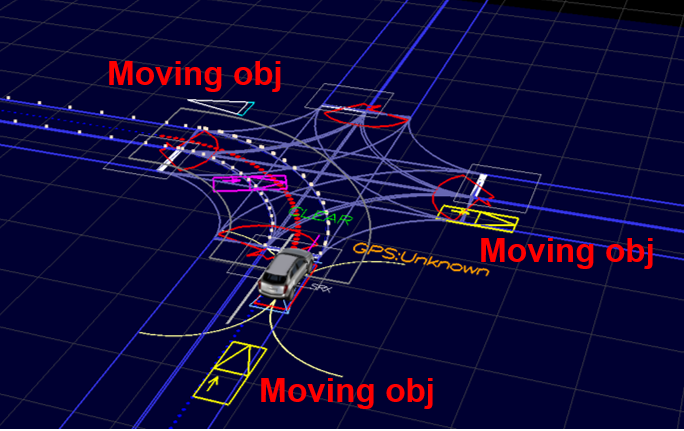}
	\hspace{7.2cm} (b) Moving Obstacles \\ in iREAD System.
          \label{fig:iread_moving}
        \end{center}
      \end{minipage}\\ \\

    \end{tabular}
    \caption{InfoRich Co-simulation Environment.}
\label{fig:overview_co-simsystems}
  \end{center}
\end{figure}

\begin{figure}[!t]
\centering
\includegraphics[width=7.55cm]{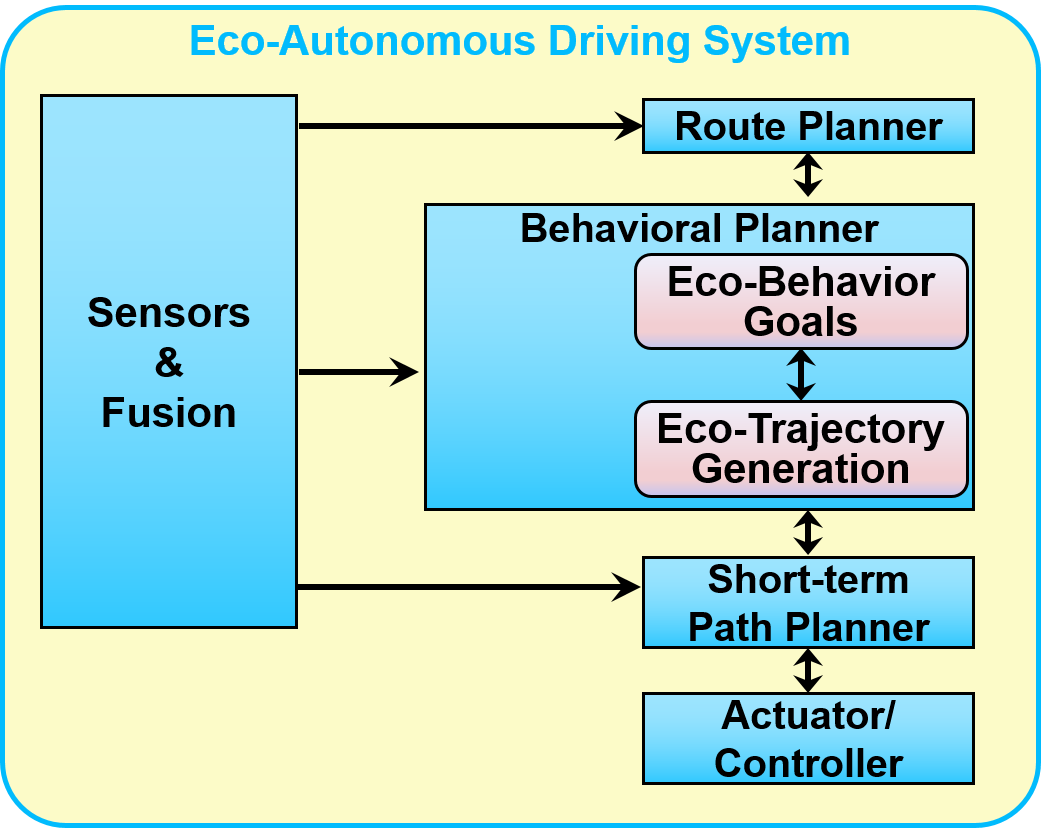}
\caption{MultiCruise Implementation on the Co-simulation Platform \cite{aoki2020inforich}.}
\label{fig:system_implementation}
\end{figure}

\begin{figure}[!t]
  \begin{center}
    \begin{tabular}{c}
      \begin{minipage}{0.5\hsize}
        \begin{center}
          \includegraphics[width=3.9cm]{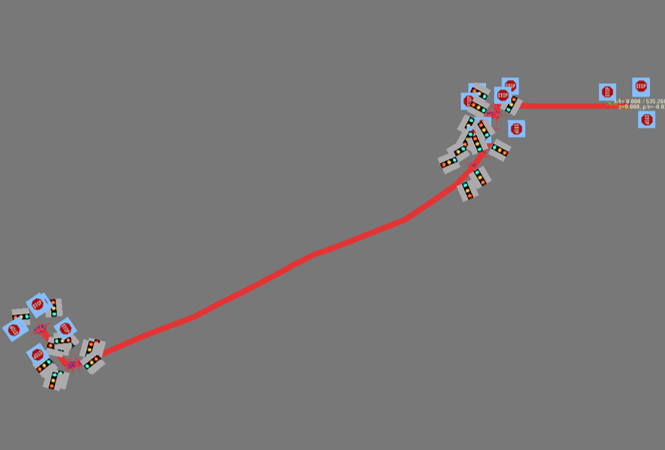}
          \hspace{6.8cm} (a) \textit{Map A}: Road networks.
          \label{fig:LongNREL_A}
          \end{center}
      \end{minipage}
      \begin{minipage}{0.5\hsize}
        \begin{center}
          \includegraphics[width=3.9cm]{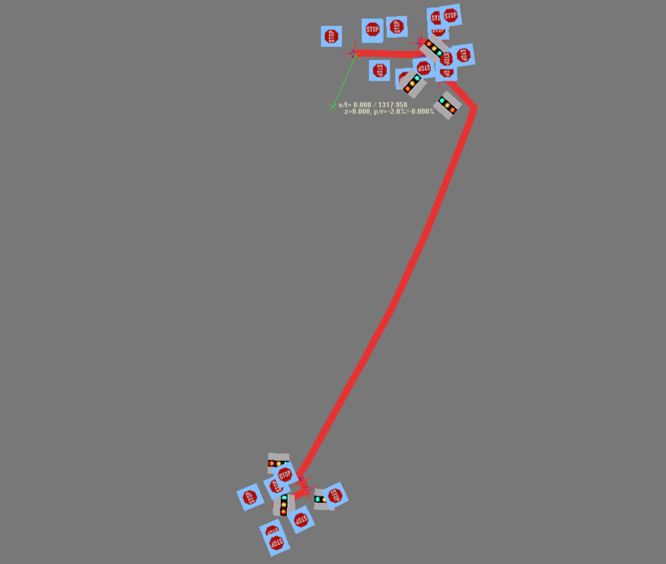}
          \hspace{6.8cm} (b) \textit{Map D}: Road networks.
          \label{fig:LongNREL_B}
          \end{center}
      \end{minipage}
    \end{tabular}
    \caption{Synthetic Drives with a Highway Road Segment.}
    \label{fig:LongTrip_NREL}
  \end{center}
\end{figure}
\section{IMPLEMENTATION AND EVALUATION}

In this section, we present the implementation and evaluation of our MultiCruise in the InfoRich co-simulation platform introduced in \cite{aoki2020inforich} that includes autonomous driving software \cite{bhat2018tools} and a Simulink-based VD\&PT model, as shown in Figure \ref{fig:harware_setup}. In the co-simulation platform, each simulator keeps exchanging the data by using the UDP protocol.
The GUIs for InfoRich co-simulation platform is presented in Figure \ref{fig:overview_co-simsystems}.
The traffic and SPaT information are generated in the traffic simulator as shown in Figure \ref{fig:overview_co-simsystems}-(a), and these information is processed to determine vehicle's behaviors in the autonomous driving software as shown in Figure \ref{fig:overview_co-simsystems}-(b).
Such co-simulation is a promising technique to model complex and dynamic systems in a distributed manner. In the co-simulation system, each subsystem runs without being aware of the entire system.
We evaluate the energy-efficient lane selection strategy in terms of fuel efficiency, and compare against a baseline protocol.
At the same time, we show the feasibility, safety and practicality of the iREAD co-simulation platform \cite{aoki2020inforich} by using it for testing and verifying our eco-driving strategy.

\begin{figure*}[!t]
  \begin{center}
    \begin{tabular}{c}
      \begin{minipage}{0.33\hsize}
        \begin{center}
          \includegraphics[width=5.25cm]{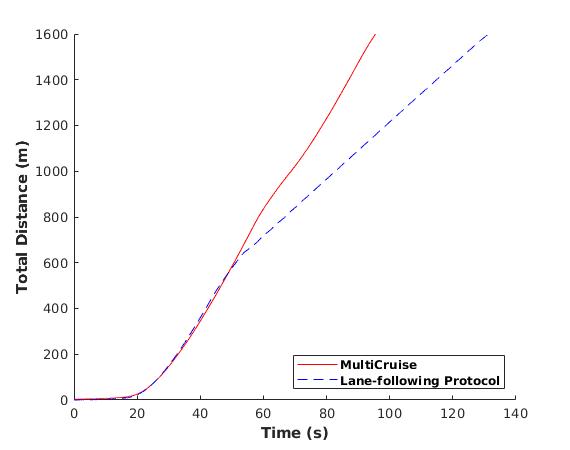}
          \hspace{7.8cm} (a) Travel Distance for Simple Scenario.
          \label{fig:SimpleDistance}
          \end{center}
      \end{minipage}
      \begin{minipage}{0.33\hsize}
        \begin{center}
          \includegraphics[width=4.75cm]{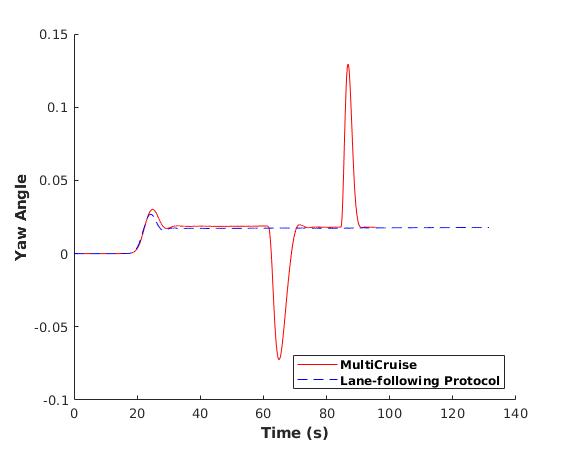}
          \hspace{6.8cm} (b) Yaw Changes for Simple Scenario
          \label{fig:SimpleYawChanges}
          \end{center}
      \end{minipage}
      \begin{minipage}{0.33\hsize}
        \begin{center}
          \includegraphics[width=4.65cm]{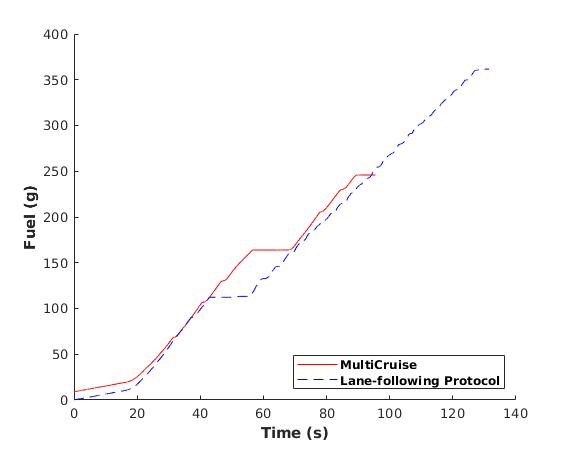}
          \hspace{6.8cm} (c) Consumed Fuel for Simple Scenario
         \label{fig:SimpleFuelTotal}
          \end{center}
      \end{minipage}
    \end{tabular}
    \caption{Performance Evaluations of MultiCruise for a Simple Scenario.}
    \label{fig:SimpleEvaluations}
  \end{center}
\end{figure*}

\begin{figure*}[!t]
\centering
\includegraphics[width=19.0cm]{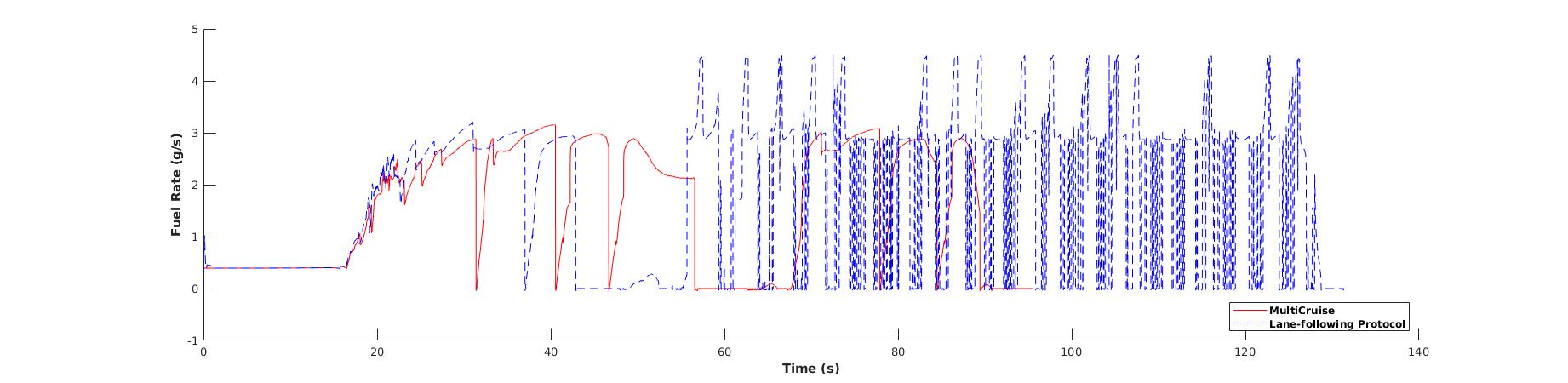}
\caption{Fuel Rates for a Simple Scenario.}
\label{fig:SimpleFuelRate}
\end{figure*}

First, we show the implementation on the eco-autonomous driving system in Figure \ref{fig:system_implementation}.
In the system, the Route Planner component calculates the route to be taken, by using the given sensor information and map information.
Based on the output, Behavioral Planner determines the lane to drive, and this component includes MultiCruise.
In Behavioral Planner, the Eco-Behavior Goals components transmit the information of the left lane, current lane and right lane, and then Eco-Trajectory Generation \cite{peter2020ecodriving} calculates the speed trajectories and cruising costs $\Gamma$ for these three lanes.
Finally, by using these three values, Eco-Behavior Goals determines the lane to drive while accounting for safety requirements.

We first evaluate our eco-driving strategy with a simple scenario in which there is only a straight road segment. In this scenario, we show the details of the vehicle movements, fuel rates, and total fuel consumption through each drive.
In addition, we evaluate MultiCruise with realistic synthetic scenarios discussed in \cite{aoki2020inforich}.
We prepare $6$ different road networks that have a variety of features, such as speed limit, road grade, curvature, and traffic controllers. These synthetic scenarios have a highway road segment, and they have more than $12$ (km) of total distance traveled.

\subsection{Performance Metric}

To evaluate our MultiCruise in terms of fuel efficiency, we use \textit{Consumed Fuel} as a metric.
The Consumed Fuel represents the fuel consumption used in each drive. The primary purpose of MultiCruise is to decrease Consumed Fuel.
The Consumed Fuel is mainly determined by vehicle dynamics and powertrain models, and the co-simulation test platform \cite{aoki2020inforich} enables to evaluate the performance.

\begin{table}[b]
\centering
\caption{Basic Information for Synthetic Cycles.}
  \begin{tabular}{c|c|c|c|c}
                          & Overall & Highway       & Speed Limit & Number of\\
                          & Distance & Distance & on Highway & Intersectons\\ \hline \hline
    \footnotesize{\textit{Map A}} & 20.31 (km) & 12.71 (km) & 105 (km/h) & 15\\ \hline 
    \footnotesize{\textit{Map B}} & 17.33 (km) & 9.86 (km) & 105 (km/h) & 10\\ \hline
    \footnotesize{\textit{Map C}} & 19.99 (km) & 4.89 (km) & 97 (km/h) & 26\\ \hline
    \footnotesize{\textit{Map D}} & 15.06 (km) & 10.79 (km) & 105 (km/h) & 8\\ \hline
    \footnotesize{\textit{Map E}} & 15.59 (km) & 3.72 (km) & 112 (km/h) & 10\\ \hline
    \footnotesize{\textit{Map F}} & 12.71 (km) & 3.46 (km) & 112 (km/h) & 14\\ 
  \end{tabular}
\label{table:NREL_highway}
\end{table}

\subsection{Synthetic Scenarios}

To show the feasibility and practicality of MultiCruise, we utilize synthetic drive cycles derived from real-world driving data provided from the National Renewable Energy Laboratory (NREL).
In particular, we select $6$ synthetic drives, \textit{Map A} to \textit{Map F}, that include highways or motorways, because MultiCruise focuses on the discretionary lane-changing maneuvers on highways.
We present $2$ examples of the synthetic cycles, as shown in Figure \ref{fig:LongTrip_NREL}.
Also, in Table \ref{table:NREL_highway}, we present the basic information of the $6$ synthetic cycles, including the cycle distance, the highway distance, the speed limit on the highway segment, and the number of road intersections.
For example, as shown in Table \ref{table:NREL_highway}, \textit{Map A} has more than $20$ (km) for overall distance and includes approximately $12$ (km) along a highway segment, in which there are no traffic controllers and/or intersections.

\subsection{Evaluation with a Simple Scenario}

We first evaluate our MultiCruise with a simple straight-road segment.
The road segment has multiple lanes to drive and has $1,600$ meters.
Also, for this case study, the speed limit is set as $112$ (km/h) that is equal to $70$ (miles/h).
We evaluate the fuel consumption, compared to a baseline \textbf{Lane-following Protocol}.
In the Lane-following Protocol, each vehicle follows the lane to drive for its navigation purpose and does not change its lane for fuel saving and/or traffic throughput.

%
%
%
%
%
%

In addition, for this case study, we have a slower lead vehicle that drives approximately at $70$ (km/h).
In the lane-following protocol, each vehicle simply drives and stays in its current lane even when there is a slower lead vehicle.

We present the traveled distance, yaw angle changes and total fuel consumption in Figures
\ref{fig:SimpleEvaluations}-(a), -(b) and -(c), respectively.
Also, we present the fuel rate in Figure \ref{fig:SimpleFuelRate}.
First, as shown in Figure \ref{fig:SimpleEvaluations}-(a), MultiCruise allows the ego vehicle to reach the destination in a much shorter trip time, because the vehicle can adaptively select the lane and start the lane-changing maneuver when it detects the slower lead vehicle.
On the other hand, the baseline protocol spends a longer trip time because of the slow lead vehicle.
Secondly, we show the yaw changes of the travels with MultiCruise and with the baseline protocol. The vehicle with MultiCruise changes its lane twice to overtake the slower lead vehicle.
On the other hand, the vehicle under the baseline protocol does not change its driving lane, and therefore, the yaw angle is constant.

Overall, MultiCruise saves around $32 \%$ fuel in this case study, as shown in Figure \ref{fig:SimpleEvaluations}-(c). The vehicle with the lane-following protocol has to slow down once it is behind the slower lead vehicle, and consequently, it wastes energy. The vehicle with MultiCruise not only does not need to hit the brakes but also can avoid wasting fuel by dynamically selecting the lane to drive at a higher speed.

\vspace{1mm}
Since this case study is with a simple scenario, we study MultiCruise in more scenarios next.

\begin{figure}[!t]
\centering
\includegraphics[width=8.00cm]{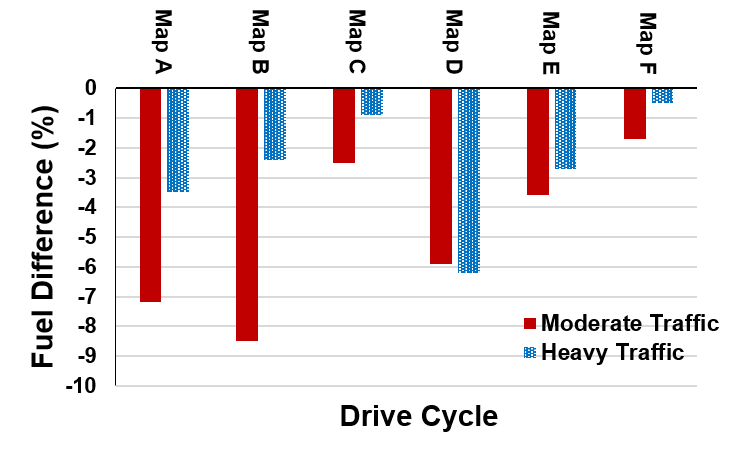}
\caption{Performance Evaluation of MultiCruise on the Fuel Consumption for Synthetic Drives.}
\label{fig:Synthetic_Fuel}
\end{figure}

\subsection{Evaluation with Synthetic Scenarios}

We evaluate MultiCruise with $6$ synthetic scenarios, \textit{Map A} to \textit{Map F}.
Each road network contains $1$ highway segment.

We compare MultiCruise to the lane-following protocol under $2$ traffic volumes: (i) Moderate traffic (100 (m/vehicle)) and (ii) Heavy traffic (50 (m/vehicle)). Since MultiCruise focuses on lane-changing maneuvers for energy efficiency, we do not evaluate the performance under the traffic jam and/or congestion.
Under the serious traffic jam, since vehicles move very slowly, the eco-lane selection strategy does not improve fuel efficiency.

As shown in Figure \ref{fig:Synthetic_Fuel}, we present the relative fuel consumption in moderate traffic and in heavy traffic.
We show the relative performance of MultiCruise, compared to the baseline protocol.
As shown in the result, MultiCruise has superior performance and successfully saves Consumed Fuel across all the cases. 
In particular, MultiCruise saves fuel consumption by $7.2 \%$ in \textit{Map A} and by $8.5 \%$ in \textit{Map B} with moderate traffic.
In \textit{Map C} and \textit{Map F}, MultiCruise has a small improvement in fuel efficiency, because the highway segment is relatively short and there are few opportunities to change lanes.
One interesting point to note is that MultiCruise saves more energy in moderate traffic than in heavy traffic, other than in \textit{Map D}.
In moderate traffic, there are more spaces and opportunities to conduct lane-changing maneuvers, and MultiCruise contributes to saving fuel consumption.
As discussed and shown in this section, MultiCruise improves fuel efficiency both in a simple scenario and in practical cases.

\section{SUMMARY AND FUTURE WORK}

In this paper, we presented a cooperative and energy-efficient lane-selection strategy named \textit{MultiCruise}, where each Connected and Automated Vehicle (CAV) selects the optimal lane and optimal target vehicle to follow from multiple candidates to save fuel consumption.
We described the Eco-Driving costs to determine whether the vehicle should stay or change the driving lane for fuel standpoint.
Also, we presented safe and cooperative mechanisms for MultiCruise that avoid selfish back-to-back lane-changing maneuvers and avoid vehicle collisions and/or accidents.
MultiCruise incorporates the Eco-Cruise component that adjusts vehicle speed and acceleration for eco-autonomous driving.
We conclude that MultiCruise improves fuel efficiency on highways and motorways by showing both simple case study and realistic simulation with the co-simulation test platform and the autonomous driving software.
In future work, we will study the energy optimization across multiple CAVs, because our work focused on the energy consumption only for the ego vehicle.
In addition, we will develop a Hardware-in-the-Loop (HIL) simulator by integrating with a real vehicle and implement MultiCruise on it.

\vspace{1mm}

\section*{Acknowledgment}
This paper is based upon the work supported by the United States Department of Energy (DOE), ARPA-E NEXTCAR program under award No. DE-AR0000797.
The authors would like to thank Jeff Gonder and Michael O'Keefe at National Renewable Energy Laboratory (NREL) for providing real world driving data and analysis.


\bibliographystyle{ieeetr}

\bibliography{IEEEabrv,../bib_From20141215}

\begin{thebibliography}{10}

\bibitem{US_EIA}
U.S. Energy Information Administration (EIA), https://www.eia.gov/.

\bibitem{wadud2016help}
Z.~Wadud, D.~MacKenzie, and P.~Leiby, ``Help or hindrance? the travel, energy
  and carbon impacts of highly automated vehicles,'' {\em Transportation
  Research Part A: Policy and Practice}, vol.~86, pp.~1--18, 2016.

\bibitem{aoki2017configurable}
S.~Aoki and R.~R. Rajkumar, ``A configurable synchronous intersection protocol
  for self-driving vehicles,'' in {\em Embedded and Real-Time Computing Systems
  and Applications (RTCSA), 2017 IEEE 23rd International Conference on},
  pp.~1--11, IEEE, 2017.

\bibitem{aoki2019DSIP}
S.~Aoki and R.~R. Rajkumar, ``V2v-based synchronous intersection protocols for
  mixed traffic of human-driven and self-driving vehicles,'' in {\em Embedded
  and Real-Time Computing Systems and Applications (RTCSA), 2019 IEEE 25th
  International Conference on}, pp.~1--11, IEEE, 2019.

\bibitem{tsugawa2016review}
S.~Tsugawa, S.~Jeschke, and S.~E. Shladover, ``A review of truck platooning
  projects for energy savings,'' {\em IEEE Transactions on Intelligent
  Vehicles}, vol.~1, no.~1, pp.~68--77, 2016.

\bibitem{peter2020ecodriving}
L.~E. Jan, J.~Zhao, S.~Aoki, A.~Bhat, C.-F. Chang, and R.~Rajkumar, ``Speed
  trajectory generation for energy-efficient connected and automated
  vehicles,'' in {\em Dynamic Systems and Control Conference}, ASME, 2020.

\bibitem{brown2014analysis}
A.~Brown, J.~Gonder, and B.~Repac, ``An analysis of possible energy impacts of
  automated vehicles,'' in {\em Road vehicle automation}, pp.~137--153,
  Springer, 2014.

\bibitem{lehmann2018generic}
B.~Lehmann, H.-J. G{\"u}nther, and L.~Wolf, ``A generic approach towards
  maneuver coordination for automated vehicles,'' in {\em 2018 21st
  International Conference on Intelligent Transportation Systems (ITSC)},
  pp.~3333--3339, IEEE, 2018.

\bibitem{hess2019negotiation}
D.~He{\ss}, R.~Lattarulo, J.~P{\'e}rez, T.~Hesse, and F.~K{\"o}ster,
  ``Negotiation of cooperative maneuvers for automated vehicles: Experimental
  results,'' in {\em 2019 IEEE Intelligent Transportation Systems Conference
  (ITSC)}, pp.~1545--1551, IEEE, 2019.

\bibitem{aoki2020inforich}
S.~Aoki, L.~E. Jan, J.~Zhao, A.~Bhat, R.~Rajkumar, and C.-F. Chang,
  ``Co-simulation platform for developing inforich energy-efficient connected
  and automated vehicles,'' in {\em 2020 IEEE Intelligent Vehicles Symposium
  (IV)}, IEEE, 2020.

\bibitem{urmson2008autonomous}
C.~Urmson, J.~Anhalt, D.~Bagnell, C.~Baker, R.~Bittner, M.~Clark, J.~Dolan,
  D.~Duggins, T.~Galatali, C.~Geyer, {\em et~al.}, ``Autonomous driving in
  urban environments: Boss and the urban challenge,'' {\em Journal of Field
  Robotics}, vol.~25, no.~8, pp.~425--466, 2008.

\bibitem{olin2019reducing}
P.~Olin, K.~Aggoune, L.~Tang, K.~Confer, J.~Kirwan, S.~R. Deshpande, S.~Gupta,
  P.~Tulpule, M.~Canova, and G.~Rizzoni, ``Reducing fuel consumption by using
  information from connected and automated vehicle modules to optimize
  propulsion system control,'' tech. rep., SAE Technical Paper, 2019.

\bibitem{oncken2020connected}
J.~Oncken, J.~Orlando, P.~K. Bhat, B.~Narodzonek, C.~Morgan, D.~Robinette,
  B.~Chen, and J.~Naber, ``A connected controls and optimization system for
  vehicle dynamics and powertrain operation on a light-duty plug-in multi-mode
  hybrid electric vehicle,'' tech. rep., SAE Technical Paper, 2020.

\bibitem{barth2019ead}
P.~{Hao}, G.~{Wu}, K.~{Boriboonsomsin}, and M.~J. {Barth}, ``Eco-approach and
  departure (ead) application for actuated signals in real-world traffic,''
  {\em IEEE Transactions on Intelligent Transportation Systems}, vol.~20,
  no.~1, pp.~30--40, 2019.

\bibitem{shao2020eco}
Y.~Shao and Z.~Sun, ``Eco-approach with traffic prediction and experimental
  validation for connected and autonomous vehicles,'' {\em IEEE Transactions on
  Intelligent Transportation Systems}, 2020.

\bibitem{jing2016design}
J.~Jing, E.~{\"O}zatay, A.~Kurt, J.~Michelini, D.~Filev, and
  {\"U}.~{\"O}zg{\"u}ner, ``Design of a fuel economy oriented vehicle
  longitudinal speed controller with optimal gear sequence,'' in {\em 2016 IEEE
  55th Conference on Decision and Control (CDC)}, pp.~1595--1601, IEEE, 2016.

\bibitem{firoozi2019safe}
R.~Firoozi, S.~Nazari, J.~Guanetti, R.~O'Gorman, and F.~Borrelli, ``Safe
  adaptive cruise control with road grade preview and v2v communication,'' in
  {\em 2019 American Control Conference (ACC)}, pp.~4448--4453, IEEE, 2019.

\bibitem{aoki2020towards}
S.~Aoki, {\em Towards Cooperative and Energy-Efficient Connected and Automated
  Vehicles}.
\newblock PhD thesis, Carnegie Mellon University, 2020.

\bibitem{NGSIM2006}
{\em Freeway Lane Selection Algorithm}.
\newblock United States Department of Transportation, Federal Highway
  Administration (FHWA), 2006.

\bibitem{jin2014improving}
Q.~Jin, G.~Wu, K.~Boriboonsomsin, and M.~Barth, ``Improving traffic operations
  using real-time optimal lane selection with connected vehicle technology,''
  in {\em 2014 IEEE Intelligent Vehicles Symposium Proceedings}, pp.~70--75,
  IEEE, 2014.

\bibitem{kang2019development}
K.~Kang, Y.~Bichiou, H.~A. Rakha, A.~Elbery, and H.~Yang, ``Development and
  testing of a connected vehicle optimal lane selection algorithm,'' in {\em
  2019 IEEE Intelligent Transportation Systems Conference (ITSC)},
  pp.~1531--1536, IEEE, 2019.

\bibitem{tian2018connected}
D.~Tian, G.~Wu, P.~Hao, K.~Boriboonsomsin, and M.~J. Barth, ``Connected
  vehicle-based lane selection assistance application,'' {\em IEEE Transactions
  on Intelligent Transportation Systems}, vol.~20, no.~7, pp.~2630--2643, 2018.

\bibitem{aoki2018dynamic}
S.~Aoki and R.~R. Rajkumar, ``Dynamic intersections and self-driving
  vehicles,'' in {\em Cyber-Physical Systems (ICCPS), 2018 IEEE/ACM 9th
  International Conference on}, pp.~320--330, ACM, 2018.

\bibitem{aoki2017merging}
S.~Aoki and R.~Rajkumar, ``A merging protocol for self-driving vehicles,'' in
  {\em Cyber-Physical Systems (ICCPS), 2017 ACM/IEEE 8th International
  Conference on}, pp.~219--228, IEEE, 2017.

\bibitem{goswami2007gap}
V.~Goswami and G.~H. Bham~PhD, ``Gap acceptance behavior in mandatory lane
  changes under congested and uncongested traffic on a multilane freeway,''
  tech. rep., 2007.

\bibitem{zhao2017analysis}
D.~Zhao, H.~Peng, K.~Nobukawa, S.~Bao, D.~J. LeBlanc, and C.~S. Pan, ``Analysis
  of mandatory and discretionary lane change behaviors for heavy trucks,'' {\em
  arXiv preprint arXiv:1707.09411}, 2017.

\bibitem{toledo2003modeling}
T.~Toledo, H.~N. Koutsopoulos, and M.~E. Ben-Akiva, ``Modeling integrated
  lane-changing behavior,'' {\em Transportation Research Record}, vol.~1857,
  no.~1, pp.~30--38, 2003.

\bibitem{dong2018continuous}
C.~Dong and J.~M. Dolan, ``Continuous behavioral prediction in lane-change for
  autonomous driving cars in dynamic environments,'' in {\em 2018 21st
  International Conference on Intelligent Transportation Systems (ITSC)},
  pp.~3706--3711, IEEE, 2018.

\bibitem{krasowski2020safe}
H.~Krasowski, X.~Wang, and M.~Althoff, ``Safe reinforcement learning for
  autonomous lane changing using set-based prediction,'' in {\em 2020 IEEE 23rd
  International Conference on Intelligent Transportation Systems (ITSC)},
  pp.~1--7, IEEE, 2020.

\bibitem{li2006realistic}
X.-G. Li, B.~Jia, Z.-Y. Gao, and R.~Jiang, ``A realistic two-lane cellular
  automata traffic model considering aggressive lane-changing behavior of fast
  vehicle,'' {\em Physica A: Statistical Mechanics and its Applications},
  vol.~367, pp.~479--486, 2006.

\bibitem{bhat2018tools}
A.~Bhat, S.~Aoki, and R.~Rajkumar, ``Tools and methodologies for autonomous
  driving systems,'' {\em PROCEEDINGS OF THE IEEE}, vol.~106, no.~9,
  pp.~0018--9219, 2018.

\end{thebibliography}

\end{document}